\documentclass[runningheads]{llncs}
\usepackage{graphicx}

\usepackage{tikz}
\usepackage{comment}
\usepackage{amsmath,amssymb} 
\usepackage{color}
\usepackage{times}
\usepackage{epsfig}
\usepackage{graphicx}
\usepackage{makecell}
\usepackage{bbding}
\usepackage{cite}
\usepackage[marginal]{footmisc}

\usepackage[accsupp]{axessibility}  

\begin{document}
\pagestyle{headings}
\mainmatter
\def\ECCVSubNumber{251}  

\title{Joint Spatial-Temporal and Appearance Modeling with Transformer for Multiple Object Tracking} 

\author{Peng Dai\inst{1} \and
Yiqiang Feng\inst{2} \and
Renliang Weng\inst{2} \and 
Changshui Zhang\inst{1}}
\institute{Tsinghua University, Beijng, China.
\email{\{daip2020, zcs\}@mail.tsinghua.edu.cn} \\
\and
Aibee Inc, Beijng, China. \email{\{yqfeng, rlweng\}@aibee.com}}
\maketitle

\footnote{Peng Dai and Yiqiang Feng contributed equally to this work.\\}

\begin{abstract}
The recent trend in multiple object tracking (MOT) is heading towards leveraging deep learning to boost the tracking performance. In this paper, we propose a novel solution named TransSTAM, which leverages Transformer to effectively model both the appearance features of each object and the spatial-temporal relationships among objects.
TransSTAM consists of two major parts: (1) The encoder utilizes the powerful self-attention mechanism of Transformer to learn discriminative features for each tracklet; (2) The decoder adopts the standard cross-attention mechanism to model the affinities between the tracklets and the detections by taking both spatial-temporal and appearance features into account.
TransSTAM has two major advantages: (1) It is solely based on the encoder-decoder architecture and enjoys a compact network design, hence being computationally efficient; (2) It can effectively learn spatial-temporal and appearance features within one model, hence achieving better tracking accuracy.
The proposed method is evaluated on multiple public benchmarks including MOT16, MOT17, and MOT20, and it achieves a clear performance improvement in both IDF1 and HOTA with respect to previous state-of-the-art approaches on all the benchmarks. Our code is available at \url{https://github.com/icicle4/TranSTAM}.
\end{abstract}

\section{Introduction}\label{section:introduction}
Multiple object tracking (MOT) in videos is an important problem in many application domains. Particularly, estimating humans location and their motion is of great interest in surveillance, business analytics, robotics and autonomous driving. 
There is a significant amount of research in this domain~\cite{braso2020learning, kim2015multiple, schulter2017deep}, and most state-of-the-art MOT works follow the tracking-by-detection (TBD) paradigm which divides the MOT task into two sub-tasks: first, obtaining frame-by-frame object detections; second, associating the set of detections into trajectories.
In this paper, we also follow the TBD strategy, and focus on the data association part.

Both the appearance features and the spatial-temporal relationships are crucial cues for MOT. Traditional data association methods~\cite{Wojke2017simple} usually rely on hand-crafted rules to fuse these affinities. Also, the spatial-temporal relationships are modeled with manually designed models~\cite{luo2017multiple}, such as linear motion model~\cite{Breitenstein2009}, social force model~\cite{FengPengming2017}, or crowd motion pattern model~\cite{HuMin2008}.
The recent trend in MOT is heading towards leveraging deep learning to boost the tracking performance.
In particular, with the success of Transformer in various computer vision tasks, such as object detection~\cite{DETR} and segmentation~\cite{strudel2021}, there are a few pioneer investigations~\cite{chu2021transmot, meinhardt2021trackformer, transtrack, xu2021transcenter} in applying Transformer to MOT. 
Both TrackFormer~\cite{meinhardt2021trackformer} and TransTrack~\cite{transtrack} are built on top of DETR architecture~\cite{DETR}, and follow the joint-detection-and-tracking framework. To mitigate the occlusion problem inherent to anchor-based bounding-box tracking methods, TransCenter~\cite{xu2021transcenter} uses dense pixel-level multi-scale queries to replace the sparse queries.
As mentioned in~\cite{chu2021transmot}, the tracking performance of the above DETR-based trackers is not satisfactory, because they fail to model the long-term spatial-temporal dependencies.
Different from~\cite{meinhardt2021trackformer, transtrack, xu2021transcenter}, Chu \emph{et al.}~\cite{chu2021transmot} proposes a spatial-temporal graph Transformer (TransMOT) model, which follows the TBD strategy and exploits the Transformer architecture to learn the affinities between the historical tracklets and the new detections.
In TransMOT, to effectively model the spatial relationship of the objects, an extra spatial graph convolutional neural network is added within the Transformer, which inevitably increases the model complexity.

\begin{figure*}[htb]
\centering
\includegraphics[width=0.90\textwidth]{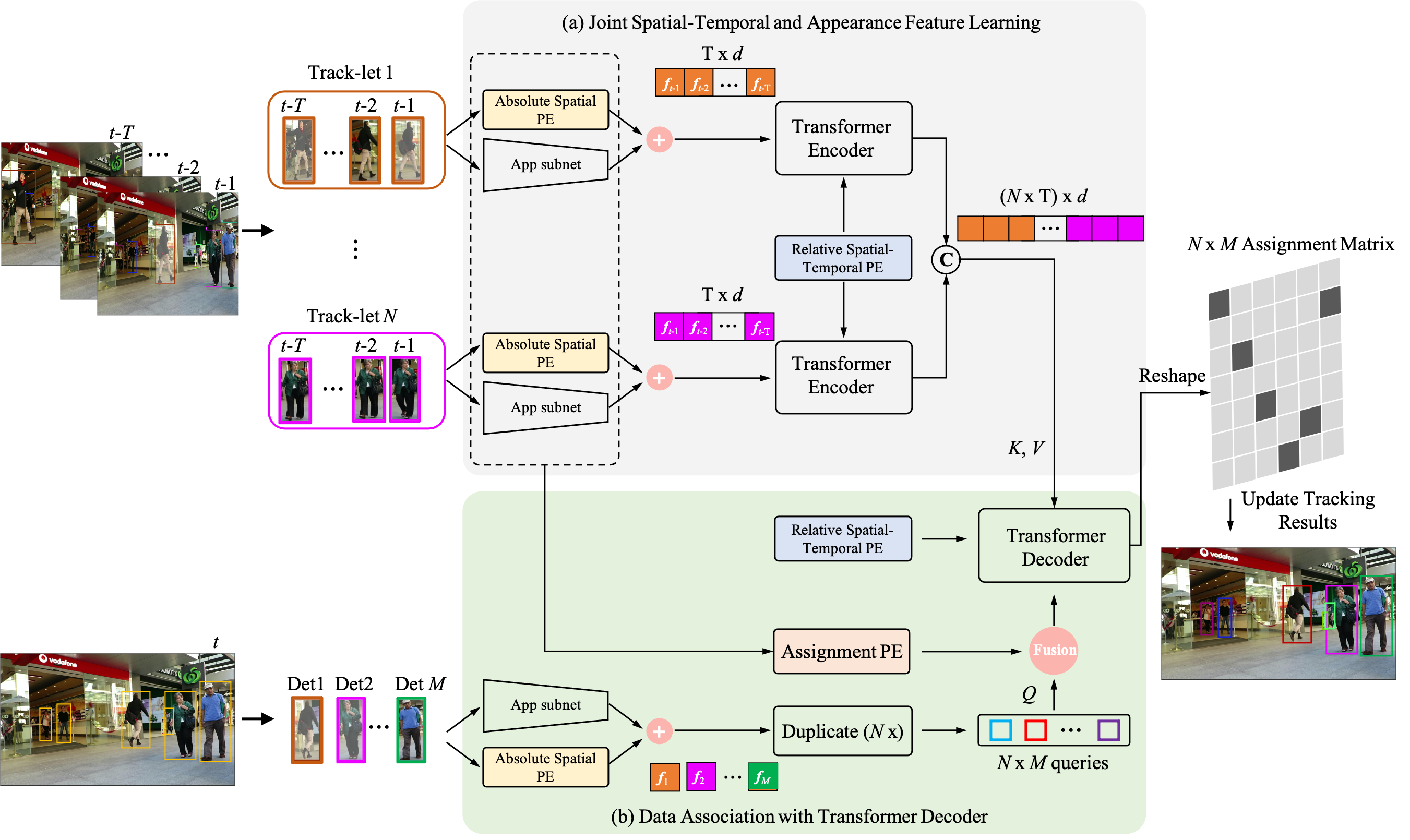}
\caption{Overview of the proposed TranSTAM for online MOT. It follows the typical encoder and decoder architecture of Transformer. Given a set of $N$ tracklets till frame $t-1$ and $M$ detections at frame $t$, a CNN model and an absolute spatial positional encoding (ASPE) model is first adopted to extract appearance and spatial features for each detection, respectively. Then, the encoder learns discriminative features for each tracklet with the help of a relative spatial temporal positional encoding (RSTPE) module. All the features are concatenated to form the memory in decoder. Next, an assignment positional encoding (APE) is proposed to generate positional encodings for $N\times M$ queries and the decoder adopts the standard cross-attention mechanism to calculate the assignment matrix $\boldsymbol{A} \in \mathbb{R}^{N \times M}$. Finally, a simple Hungarian algorithm is adopted to generate the association results.}
\label{fig:picture001}
\end{figure*}

In this paper, we propose a simple yet effective solution named TransSTAM to solve the MOT problem with Transformer. Similar to TransMOT~\cite{chu2021transmot}, our method also follows the TBD strategy and utilizes Transformer to learn the affinities between the tracked objects and detections in the current frame.
Moreover, our method is solely based on the encoder-decoder architecture, and this compact network design enjoys good performance with less computational cost compared to TransMOT.
To apply Transformer in MOT, we propose three simple yet effective positional encoding methods.
First, we propose an \textit{Absolute Spatial Positional Encoding} (ASPE) and a \textit{Relative Spatial-Temporal Positional Encoding} (RSTPE) method for representation learning.
As shown in Figure~\ref{fig:picture001}, ASPE seeks to obtain a representation that encodes the bounding box coordinates of each detection, and RSTPE captures the relative spatial-temporal relation among detections.
Equipped with these encodings, Transformer can better model the spatial-temporal relationships of detections and effectively integrate them with the appearance features.
Second, we propose an \textit{Assignment Positional Encoding} (APE) for affinity modeling.
With APE, the encoded features of all tracklets can be used when calculating the affinity between certain detection-tracklet pair in decoder, which enlarges the query's field of view to the global context and thereby improving final tracking performance.

The main contribution of the paper is in three folds: 
(1) We propose a novel Transformer based method named TranSTAM for MOT, which enjoys a compact network design and is computationally efficient. 
(2) We propose three simple yet effective positional encoding methods on the basis of the Transformer for representation learning and affinity modeling. 
(3) We show significantly improved state-of-the-art results of our method on multiple MOTChallenge benchmarks.

\section{Related Work}\label{section:relatedwork}
Most state-of-the-art MOT trackers follow the TBD paradigm.
The TBD framework generates tracklets by associating detections on a frame-by-frame basis for online applications~\cite{hu2020multi, Wojke2017simple, xu2019spatial, zhou2018deep, zhu2018online} or a batch basis for offline scenarios~\cite{berclaz2011multiple, braso2020learning, milan2015multi}.
Traditional data-association methods differ in the specific optimization methods, including network flow \cite{Pirsiavash2011Globally}, generalized maximum multi clique \cite{dehghan2015gmmcp}, linear programming \cite{jiang2007linear}, conditional random field \cite{yang2012online}, \emph{etc}.
However, the authors in \cite{bergmann2019tracking} showed that the significantly higher computational cost of these over-complicated optimization methods does not translate to significantly higher accuracy.

Recently, deep learning-based association algorithm is gaining popularity in MOT~\cite{braso2020learning, chu2019famnet, peng2021lpc, schulter2017deep, xu2020train}. 
Chu \emph{et al.}~\cite{chu2019famnet} proposed an end-to-end model, named FAMNet, to refine feature representation, affinity model and multi-dimensional assignment in a single deep network.
Xu \emph{et al.}~\cite{xu2020train} presented a differentiable Deep Hungarian Net (DHN) to approximate the Hungarian matching algorithm and provided a soft approximation of the optimal prediction-to-ground-truth assignment.
Schulter \emph{et al.}~\cite{schulter2017deep} designed a bi-level optimization framework which frames the optimization of a smoothed network flow problem as a differentiable function of the pairwise association costs.
Bras\'o \emph{et al.}~\cite{braso2020learning} modeled the non-learnable data-association problem as a differentiable edge classification task.
Dai \emph{et al.}~\cite{peng2021lpc} proposed a proposal-based learnable framework, which is similar to the two-stage object detector Faster RCNN and models MOT as a proposal generation, proposal scoring and trajectory inference paradigm. The proposal scoring can be solved by a learnable graph convolutional network.

Recently, Transformer based architectures are applied to various tasks such as object detection~\cite{DETR} and segmentation~\cite{strudel2021}.
There are also a few pioneer investigations~\cite{chu2021transmot, meinhardt2021trackformer, transtrack, xu2021transcenter} in applying Transformer to MOT. 
Tim \emph{et al.}~\cite{meinhardt2021trackformer} and Sun \emph{et al.}~\cite{transtrack} built the Transformer-based trackers on DETR architecture~\cite{DETR}, and followed the joint-detection-and-tracking framework. One common feature of these two methods is that the data association is achieved with attention operations over feature map of the current frame and track queries from the previous frames.
To mitigate the occlusion problem inherent to anchor-based bounding-box tracking methods, Xu \emph{et al.}~\cite{xu2021transcenter} replaced the sparse queries with dense pixel-level multi-scale queries.
As mentioned in~\cite{chu2021transmot}, the tracking performance of the above DETR-based trackers is not the state-of-the-art, because they cannot model the long-term spatial-temporal dependencies.
Different from~\cite{meinhardt2021trackformer, transtrack, xu2021transcenter}, Chu \emph{et al.}~\cite{chu2021transmot} proposed a spatial-temporal graph Transformer (TransMOT) model, which follows the TBD strategy and exploits the Transformer architecture to learn the affinities between the historical tracklets and the new detections.
In TransMOT, an extra spatial graph convolutional neural network is added within the Transformer to model the spatial relationship of the objects at each timestamp independently, which may limit the Transformer's ability in modeling global attention and be less computationally efficient.
Meanwhile, due to the lack of Positional Encoding (PE) in the decoder, the affinity estimation of each detection-tracklet pair can only rely on the local features within this tracklet.
It may reduce the ability of capturing long-range dependencies among all tracklets through cross-attention, which is arguably the main source for the success of Transformer.

\section{Method}\label{section:method}
\subsection{Framework Overview}\label{subsectionFO}
As shown in Figure~\ref{fig:picture001}, our framework is built upon tracking-by-detection framework, and aims at tracking multiple objects in an online fashion.
Assume that, at time step $t$, the framework maintains a set of $N$ tracklets $\Xi =\{\mathcal{T}_{1}, \cdots ,\mathcal{T}_{N}\}$, each of which represents a tracked object and is generated by linking detections over the previous $T$ image frames. 
It should be noticed that not every tracklet has all the detections in the previous $T$ frames due to occlusion, missing detection, \emph{etc}.
For illustration purpose, we consider the situation that there is no missing detections for each tracklet. The extension of our method to the cases with missing detections is easy and  straightforward.
Meanwhile, a set of $M$ detections $\mathcal{D}=\{\boldsymbol{d}_{1}, \cdots ,\boldsymbol{d}_{M}\}$ is obtained by applying an object detector on frame $t$.
The task of online MOT is to associate detections for the existing tracklets, determine whether any tracklets should be terminated, and generate new tracklets for new objects that enter the scene.

As shown in Figure~\ref{fig:picture001}, our framework consists of two major parts:
(1) utilize the powerful self-attention mechanism of Transformer encoder to jointly model the spatial-temporal information and fuse with the appearance feature for each tracklet;
(2) calculate the $N\times M$ assignment matrix with the help of the standard multi-head self- and cross-attention mechanisms in Transformer decoder.

\begin{figure}[tb]
\centering
\includegraphics[width=0.49\textwidth]{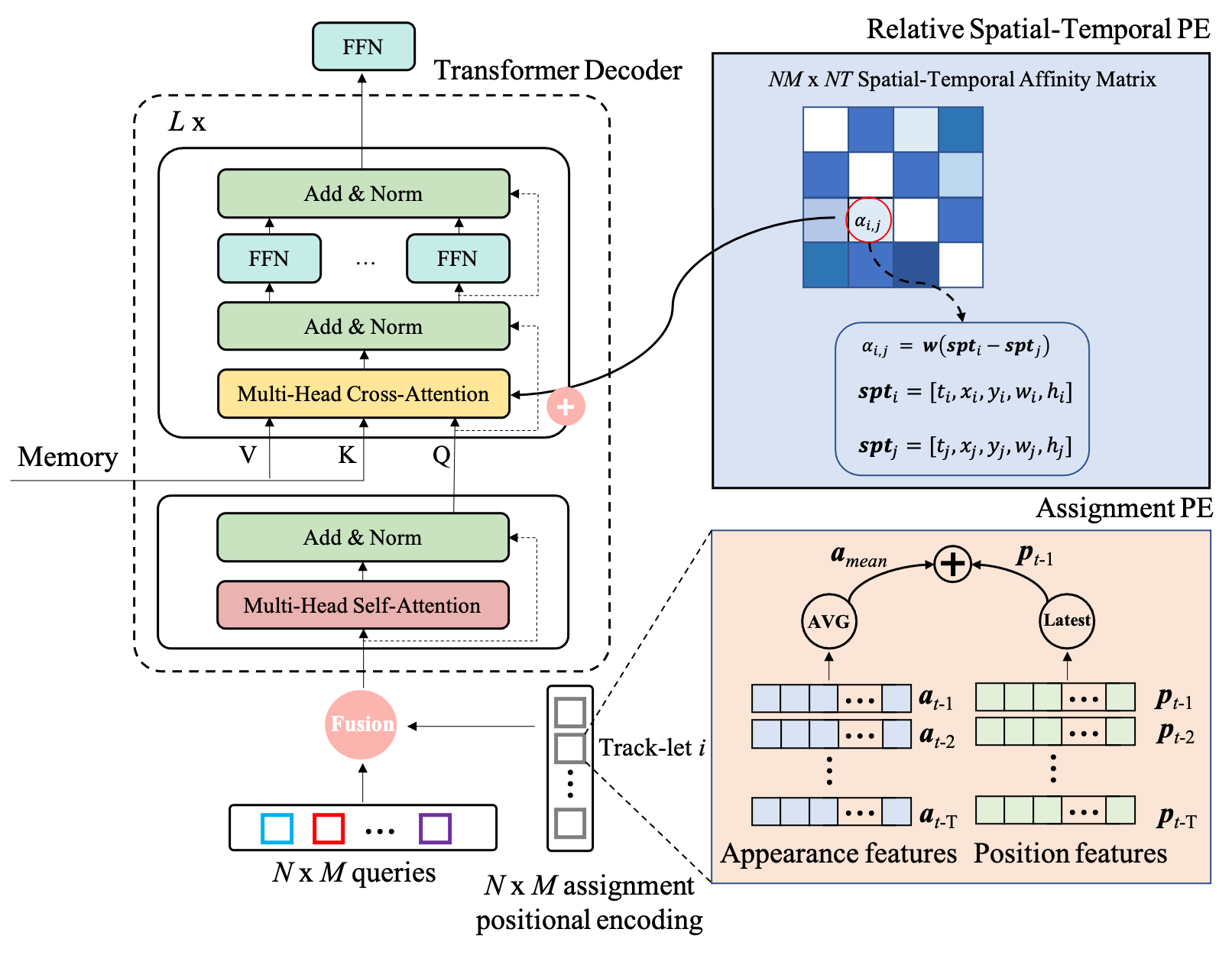}
\caption{Architecture of spatial-temporal transformer decoder.}
\label{fig:picture003}
\end{figure}

In the first part, a CNN with shared weights is used to extract appearance feature $\boldsymbol{a}_{i} \in \mathbb{R}^{d}$ directly from RGB data of each detection $\boldsymbol{d}_{i}$.
To model the spatial-temporal information of each tracklet more effectively, we decompose the spatial-temporal modeling into absolute spatial modeling of each detection and relative spatial-temporal modeling between detections.
For absolute spatial modeling, we introduce a ASPE, a 3-layer MLP with shared weights, to extract feature embeddings $\boldsymbol{p}_{i} \in \mathbb{R}^{d}$ on the normalized bounding box coordinates ($\Bar{x}_{i}, \Bar{y}_{i}, \Bar{w}_{i}, \Bar{h}_{i})$ of $\boldsymbol{d}_{i}$.
Then, for each detection $\boldsymbol{d}_{i}$ and each tracklet $\mathcal{T}_{i}$, its feature embedding can be represented as $\boldsymbol{f}_{i}=\boldsymbol{a}_{i} + \boldsymbol{p}_{i}$ and $\mathcal{T}_{i}=[\boldsymbol{f}_{j}]^{t-1}_{j=t-T}$, respectively.
For relative spatial-temporal modeling, as shown in Figure~\ref{fig:picture002} and Figure~\ref{fig:picture003}, we propose a RSTPE to capture the spatial-temporal relationships between detections and then use it as a bias term in the self- and cross-attention module.
Concretely, for each detection pair, RSTPE computes the dot-product of the relative spatial-temporal features and the learnable embeddings.
Incorporating with the spatial information obtained by ASPE and a better attention mechanism with RSTPE, the Transformer encoder could learn a more discriminative feature representation for each tracklet.
The details of joint spatial-temporal and appearance feature learning with Transformer encoder will be explained in Sec.~\ref{subsectionEncoder}.

In the second part, we aim at computing an assignment matrix $\boldsymbol{A} \in \mathbb{R}^{N \times M}$ between the existing $N$ tracklets and $M$ detections with the help of Transformer decoder.
Two ingredients are essential for generating the assignment matrix $\boldsymbol{A}$: 
(1) an APE that forces the order of the predicted assignment results;  
(2) an RSTPE that captures the spatial-temporal relationships between tracklets and detections.
As shown in Figure~\ref{fig:picture003}, for $(i,j)$-th query $\boldsymbol{q}_{ij}$, its query feature is represented as the feature embedding of $j$-th detection, and its corresponding positional encoding is extracted from $i$-th tracklet with APE.
By fusing the query feature with its corresponding positional encoding, we can make $A_{ij}$ (\emph{i.e.}, $(i,j)$-th element of $\boldsymbol{A}$) explictly correspond to the affinity between tracklet $\mathcal{T}_{i}$ and detection $\boldsymbol{d}_{j}$.
Meanwhile, with APE, each query detection can attend to all tracklets, hence taking full advantage of Transformer's global receptive field to improve the association accuracy.
In Transformer decoder, RSTPE is utilized in the cross-attention module to encode the relative spatial-temporal relationships between tracklets and detections.
Different from TransMOT~\cite{chu2021transmot} which first models the spatial relationship of different objects at each timestamp independently and then encodes the temporal dimension for each tracklet independently, RSTPE could encode both spatial and temporal relationship of any two detections simultaneously. It makes our model more accurate.
The details of data association with Transformer decoder will be explained in Sec.~\ref{subsectionDecoder}.

\subsection{Feature Learning with Transformer Encoder}\label{subsectionEncoder}

\begin{figure}[tb]
\centering
\includegraphics[width=0.49\textwidth]{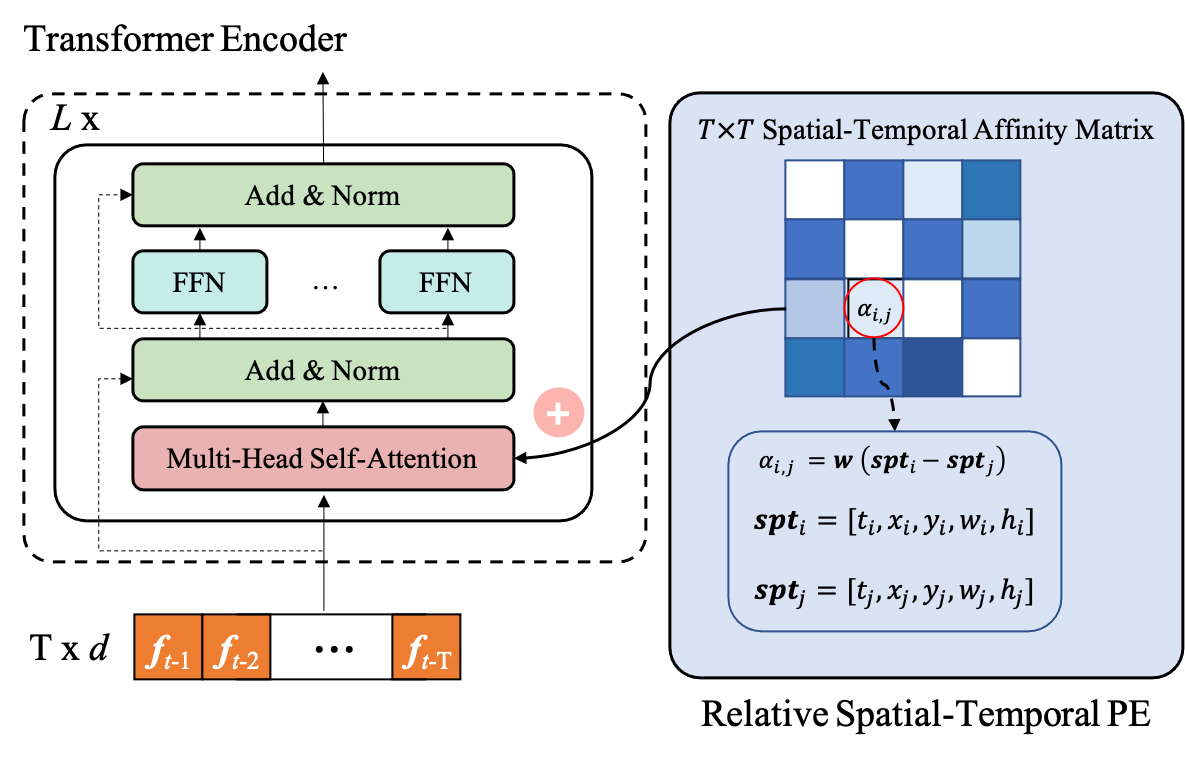}
\caption{Architecture of spatial-temporal transformer encoder.}
\label{fig:picture002}
\end{figure}

The Transformer encoder aims at learning a more discriminative feature representation for each tracklet.
Most of the existing Transformer-based trackers~\cite{meinhardt2021trackformer, transtrack, xu2021transcenter} heavily rely on the appearance features, and cannot model long term spatial-temporal dependencies. 
In this paper, we will simultaneously model the appearance and the spatial-temporal features with one Transformer model.
The input of Transformer encoder is a set of tracklets $\Xi =\{\mathcal{T}_{1}, \cdots ,\mathcal{T}_{N}\}$ for the past $T$ frames.
Since all the tracklets are processed by the encoder independently, we take one tracklet $\mathcal{T}_{i} = \{\boldsymbol{d}^{i}_{t-1}, \cdots ,\boldsymbol{d}^{i}_{t-T}\}$ as an example for illustration. Without introducing ambiguity, we  omit index $i$ to ease reading in the following, \emph{i.e.}, $\mathcal{T} = \{\boldsymbol{d}_{t-1}, \cdots ,\boldsymbol{d}_{t-T}\}$.

As described in Sec.~\ref{subsectionFO}, the appearance and spatial features for each detection are first embedded through a CNN and an MLP network, respectively.
Then, a simple ``add'' fusion operator is adopted to fuse the appearance and spatial features for each detection. These detection features are arranged into a feature tensor $\mathcal{F}=[\boldsymbol{f}_{t-1}^T, \cdots, \boldsymbol{f}_{t-T}^T]^T \in \mathbb{R}^{T \times d}$, where $d$ is the output dimension of the embedding layer.
It is further passed to the Transformer encoder module, as shown in Figure~\ref{fig:picture002}.
The Transformer encoder module consists of $L \times$ Transformer layers~\cite{vaswani2017attention}, where each Transformer layer has two parts: a self-attention module and a feed-forward
network (FFN).
The standard self-attention is calculated as:
\begin{equation}\label{equa:self-attention1}
    Q=\mathcal{F} W_{Q}, \quad K=\mathcal{F} W_{K}, \quad V=\mathcal{F} W_{V}
\end{equation}
\begin{equation}\label{equa:self-attention2}
    A^{attn}=\frac{QK^T}{\sqrt{d_{k}}}, \quad Attention(\mathcal{F}) = softmax(A^{attn})V
\end{equation}
where $W_{Q}\in \mathbb{R}^{d \times d_{k}}$, $W_{K}\in \mathbb{R}^{d \times d_{k}}$, $W_{V}\in \mathbb{R}^{d \times d_{v}}$ are projection matrices, $d_{k}$ and $d_{v}$ indicate the scaling factors, $A^{attn}$ is a matrix capturing the similarity between queries and keys.
For simplicity of illustration, we consider the single-head self-attention and assume $d_{k}=d_{v}=d$. The extension to the multi-head attention is standard and straightforward.

To encode the relative spatial-temporal information between detections, we propose RSTPE.
As shown in Figure~\ref{fig:picture002}, for each detection pair $(\boldsymbol{d}_{i}, \boldsymbol{d}_{j})$, we utilize the relative spatial-temporal feature $[\delta t, \delta x, \delta y, \delta w, \delta h]^T$ as its edge feature and compute the dot-product of the edge feature and a learnable embedding.
Then, the dot-product is used as a bias term to the attention module.
Concretely, we modify the $(i, j)$-element of $A^{attn}$ in Eq.~\ref{equa:self-attention2} with the relative spatial-temporal encoding $a_{i,j}$ as:
\begin{equation}\label{equa:RSTPE1}
    A_{i,j}^{attn}=\frac{(\boldsymbol{f}_{i} W_{Q})(\boldsymbol{f}_{j} W_{K})^{T}}{\sqrt{d}} + a_{i,j} 
\end{equation}
\begin{equation}\label{equa:RSTPE2}
    a_{i,j}= \boldsymbol{w} \cdot [\delta t, \delta x, \delta y, \delta w, \delta h]^T
\end{equation}
where $\boldsymbol{w}$ represents the learnable embedding, and shared across all layers.

The attention weighted feature tensor is projected through a FFN and a normalization layer. The features of all the tracklets are concatenated to get the final output of the Transformer encoder layer $\mathcal{F}^{en} \in \mathbb{R}^{NT \times d}$.

\subsection{Data Association with Transformer Decoder}\label{subsectionDecoder}

The Transformer decoder generates the assignment matrix $\boldsymbol{A}$ from the output of the Transformer encoder $\mathcal{F}^{en}$ and the features of $M$ detections $\mathcal{F}^{det} \in \mathbb{R}^{M \times d}$.
First, $\mathcal{F}^{det}$ is duplicated $N$ times such that $\mathcal{F}^{det} \to \mathcal{F}^{det'} \in \mathbb{R}^{NM \times d}$.
Then, $\mathcal{F}^{det'}$ is utlized as the queries and is passed into the Transformer decoder module.

Since the decoder is permutation-invariant, the $N\times M$ queries must be different to produce different results.
As mentioned above, when estimating the elements of the same column (\emph{e.g.}, $i$-th column) in $\boldsymbol{A}$,  their corresponding query features are just the same (the feature of $i$-th detection).
In order to solve this issue, we propose our APE.
For $(i, j)$-th query which is used to calculate the affinity between $i$-th tracklet and $j$-th detection, APE extracts feature $\boldsymbol{f}_{i}^{PE}$ from $i$-th tracklet as positional encoding and then fuse it with the feature embedding $\boldsymbol{f}_{j}$ of $j$-th detection to form the final feature $\boldsymbol{q}_{ij}$.
\begin{equation}\label{equa:APE1}
    \boldsymbol{f}_{i}^{PE} = \frac{1}{T} \sum_{k=1}^{T} \boldsymbol{a}_{t-k} + \boldsymbol{p}_{t-1}
\end{equation}
\begin{equation}\label{equa:APE2}
    \boldsymbol{q}_{ij} = \phi (\boldsymbol{f}_{j}, \boldsymbol{f}_{i}^{PE})
\end{equation}
where $\{\boldsymbol{a}_{t-1}, \cdots \boldsymbol{a}_{t-T}\}$ indicates the appearance features of $i$-th tracklet in the past $T$ frames, $\boldsymbol{p}_{t-1}$ represents the latest spatial feature of $i$-th tracklet, $\phi$ is a fusion function.
As for the feature fusion $\phi$, we have tried three different fusion operators: 
(1) add the query feature and positional encoding together;
(2) concatenate the query feature and positional encoding;
(3) subtract positional encoding from the query feature.
We will show how different fusion operator affects the performance in the experiments. 
The ``subtract'' fusion operator is adopted in our final network for its best performance.

The decoder takes the detection features of the current frame $\mathcal{F}^{det'} \in \mathbb{R}^{NM \times d}$ as the query, and uses the encoded features of all the tracklets $\mathcal{F}^{en} \in \mathbb{R}^{NT \times d}$ as the key and value.
The decoder follows the standard architecture of the Transformer~\cite{DETR}, transforming $N\times M$ embeddings of size $d$ using multi-head self- and cross-attention mechanisms.
The difference with the original Transformer~\cite{DETR} is that our model utilizes RSTPE to encode the relative spatial-temporal distance between the query detection and the tracklet detection, and serves it as a bias term in the cross-attention module.
By using RSTPE, each query in a single Transformer layer can adaptively attend to all tracklet detections according to the learned attention weights. 
For example, if $a_{i,j}$ in Eq.~\ref{equa:RSTPE1} is learned to be a decreasing function with respect to spatial distance. For each query, the model will likely pay more attention to the tracklet detections near it and pay less attention to the tracklet detections far away from it.
One benefit of using Transformer decoder to calculate the assignment matrix is that the Transformer layer could provide a global receptive field. 
Therefore, all the tracklet features can be utilized when calculating the affinity of specific detection-tracklet pair, which is critical to avoid ID switches in MOT.

The output of the Transformer decoder $\mathcal{F}^{de} \in \mathbb{R}^{NM \times d}$ can be passed through a FFN and a softmax layer to generate the assignment matrix $\boldsymbol{A} \in \mathbb{R}^{N \times M}$.

\subsection{Training and Inference}\label{subsectionTraining}
The proposed model is trained end-to-end with the guidance of the groundtruth assignment matrix $\boldsymbol{A}^{g}$.
We formulate the prediction of the assignment matrix as a binary classification problem. Therefore, the cross-entropy loss is applied to optimize the network.
\begin{equation}\label{equa:loss}
\begin{aligned}
    \mathcal{L} = \frac{-1}{MN}\sum_{i=1}^{N}\sum_{j=1}^{M} & A_{i, j}^{g}log(A_{i, j}) \ + \\ & (1 - A_{i, j}^{g})\log{(1-A_{i, j})}
\end{aligned}
\end{equation}
where $A_{i, j}$ and $A_{i, j}^{g}$ indicate the $(i,j)$-th element of the predicted assignment matrix $\boldsymbol{A}$ and the groundtruth assignment matrix $\boldsymbol{A}^{g}$, respectively.
In MOT, each detection in frame $t$ can only have either one matched tracklet or no match at all. In other words, each row and column of the $\boldsymbol{A}^{g}$ can only be a one-hot vector (\emph{i.e.}, a vector with 1 in a single entry and 0 in all other entries) or an all-zero vector. It will cause serious data imbalance issue.
To mitigate the data imbalance, we will downsample negative pairs and try to keep the number of positive and negative pairs comparable.

In each training iteration, a continuous sequence of $T$ +1 frames are randomly sampled from the training set. The bounding boxes and their corresponding IDs are collected from each frame.
Similar to~\cite{chu2021transmot}, the groundtruth bounding boxes are replaced by the bounding boxes generated from the object detector by matching their IoUs, hence making the model be more robust to detection noise.
We do data augmentation by randomly removing detections from tracklets to simulate missing detections.

During inference, our model first calculates the assignment matrix $\boldsymbol{A}$.
To reduce the computational cost and accelerate the inference, a simple filtering strategy based on the spatial-temporal distances is adopted to reduce the number of queries. Precisely, all the tracklet-detection pairs whose average moving speed is larger than a threshold will be dropped.
Then, a threshold $\tau_{th}$ is adopted on $\boldsymbol{A}$ for binarization. 
In other words, only the tracklet-detection pairs with affinity larger than $\tau_{th}$ can be associated.
Next, a simple Hungarian algorithm~\cite{MunkresJames} is adopted to generate tracking output, while complying with the typical tracking constraints like no detection assigned to more than one tracklet.
The detections that do not match any tracklet will be assigned with a new ID, and the tracklets that do not match any detections in the past consecutive $T$ frames will be terminated.

\section{Experiment}
\subsection{Experimental Setup}
\subsubsection{Datasets and Metrics}
\ 
\newline
\indent 
All experiments are done on the multiple object tracking benchmark MOTChallenge, which consists of several challenging pedestrian tracking sequences with frequent occlusions and crowded scenes.
We choose three separate tracking benchmarks, namely MOT16~\cite{milan2016mot16}, MOT17~\cite{milan2016mot16} and MOT20~\cite{dendorfer2020mot20}.
These three benchmarks consist of challenging video sequences with varying viewing angle, size, number of objects, camera motion, illumination and frame rate in unconstrained environments.
To ensure a fair comparison with other methods, we use the public detections provided by MOTChallenge, and regress them by using the same method as in~\cite{Stadler_2021_CVPR}. 
This regression strategy is widely used in published methods~\cite{braso2020learning, inproceedingsLiu, Stadler_2021_CVPR}.

For the performance evaluation, we use the widely accepted MOT metrics~\cite{bernardin2008evaluating, 2020HOTA, ristani2016performance, wu2006tracking}, including Multiple Object Tracking Accuracy (MOTA), ID F1 score (IDF1), Higher Order Tracking Accuracy (HOTA), Mostly Track targets (MT), Mostly Lost targets (ML), False Positives (FP), False Negatives (FN), ID switches (IDs), \emph{etc}.
Among these metrics, MOTA, IDF1 and HOTA are the most important ones. 
MOTA and IDF1 quantify two of the main aspects of MOT, namely, detection and association. 
HOTA balances the effect of performing accurate detection and association into a single unified metric.
IDF1 and HOTA are preferred over MOTA for evaluation due to their ability in measuring association accuracy.

\subsubsection{Implementation Details}
\ 
\newline
\indent 
\textbf{ReID Model.} 
Similar to LPC~\cite{peng2021lpc}, we also employ a variant of ResNet50, named ResNet50-IBN~\cite{luo2019strong}, to extract ReID features.
And, the ResNet50-IBN model is trained on two publicly available datasets: ImageNet~\cite{deng2009imagenet} and Market1501~\cite{zheng2015scalable}.

\textbf{Parameter Setting.} 
Both the encoder and decoder apply 2 individual layers of feature width 256. Each attention layer applies multi-headed self-attention~\cite{vaswani2017attention} with 8 attention heads. 
The feature dimension of FFN is set to 1024.
We set the length of the temporal window to $T=150$, which means that the tracklets without merged to any detections for 150 frames will be terminated.

\textbf{Training.}  
The transformer model is trained end-to-end with SGD optimizer, where weight decay term is set to $1 \times 10^{-4}$ and $\beta_{1}$  is set to 0.9, respectively.
The batch size is set to 4.
We train for 10 iterations in total with a learning rate $1 \times 10^{-3}$. 

\textbf{Post Processing.} 
We perform simple bilinear interpolation along missing frames to fill gaps in our trajectories.

\subsection{Ablation Study}
The ablation experiments are evaluated on the training sequences of the MOT17 dataset with a 3-fold cross-validation split as described in~\cite{braso2020learning}.

\textbf{Importance of ASPE and RSTPE.}
There are two kinds of positional encodings in feature learning: ASPE and RSTPE. In order to validate their effectiveness, we perform experiments with various combinations of positional encodings. And the results can be found in Table~\ref{table:abs_sptmha}.
It can be concluded that: (i) both ASPE and RSTPE play an important role in identity preservation, hence improving IDF1 score by 0.7 and 0.8 respectively, (ii) combination of these two positional encodings can further improve the tracking performance.
Given these ablations, we conclude that both ASPE and RSTPE play an important role in the final tracking performance.

\begin{table}[tp]
\begin{center}
\caption{Results for different combinations of positional encoding in feature learning.}
\label{table:abs_sptmha}
\begin{tabular}{p{1.25cm}<{\centering} p{1.25cm}<{\centering} c c}
\hline
ASPE & RSTPE & IDF1 & MOTA \\
\hline
 &  & 67.7 & 62.7\\
\checkmark &  & 68.4 & 62.8 \\
 & \checkmark & 68.5 & 62.8 \\
\checkmark & \checkmark & 69.2 & 63.2 \\  
\hline
\end{tabular}
\end{center}
\end{table}

\begin{figure*}[htp]
\centering
\includegraphics[width=0.90\textwidth]{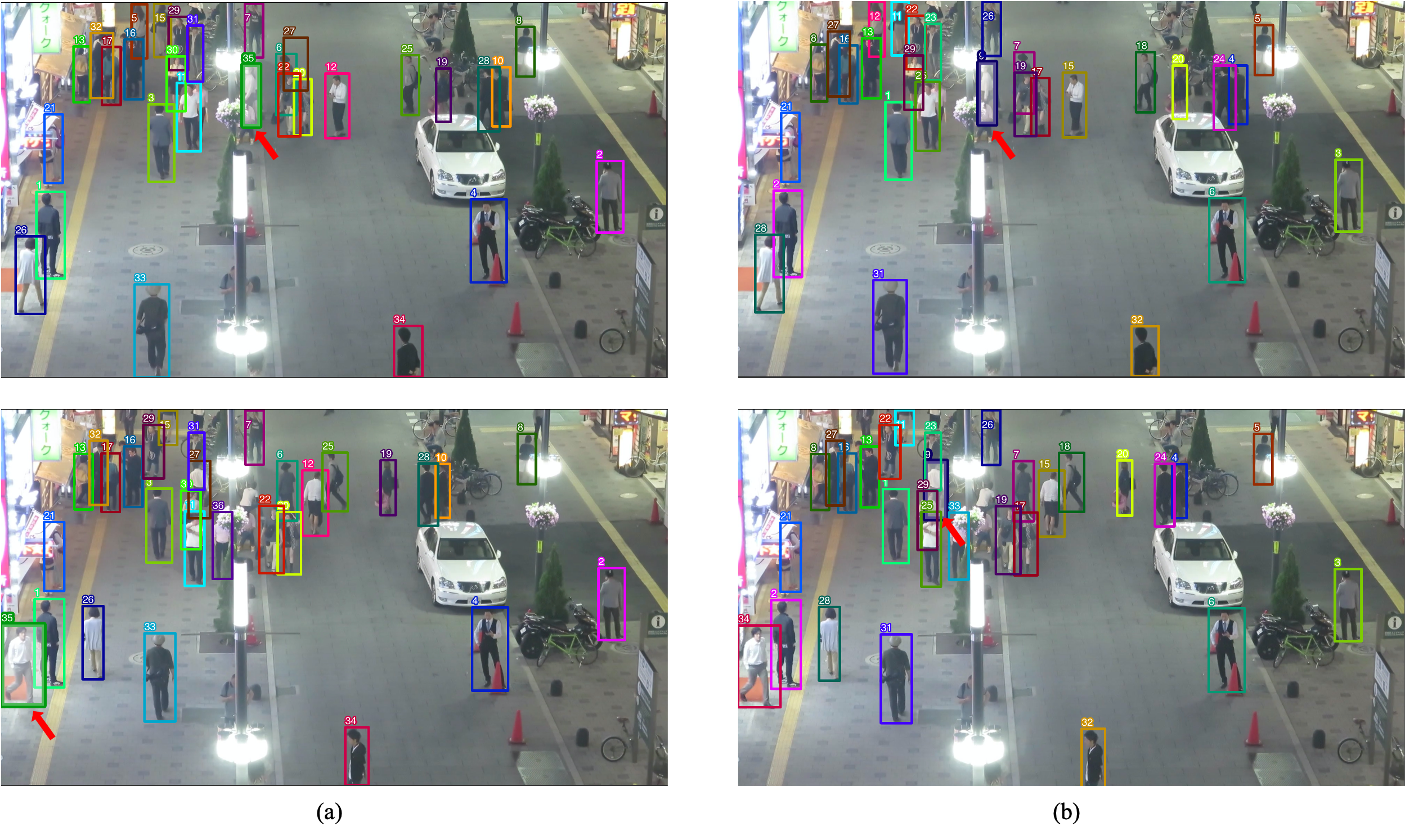}
\caption{A qualitative example showing a failure case, as shown in column (a), without using ASPE and RSTPE, which leads to a spatial-temporal jumping case when one person is fully occluded. Using ASPE and RSTPE can effectively handle this case, as shown in column (b). The numbers are the object IDs. Best viewed in color.}
\label{fig:picture004}
\end{figure*}

In order to further illustrate the effectiveness of ASPE and RSTPE, we also give the qualitative results with and without ASPE and RSTPE. As shown in Figure~\ref{fig:picture004} (a), without using ASPE and RSTPE, there is an abnormal identity transfer when the person with ID 35 is fully occluded. By introducing ASPE and RSTPE, this case can be solved, as shown in Figure~\ref{fig:picture004} (b). It demonstrates that ASPE and RSTPE can effectively model the spatial-temporal information, and improve the identity preservation in occlusion cases.

\textbf{Effectiveness of Fusion Operators in Decoder.}
As discussed in Section~\ref{subsectionDecoder}, a fusion operator is needed in Transformer decoder to fuse the query feature and its corresponding positional encoding.
We perform an experiment to study the impact of different fusion operators. Table~\ref{table:abs_DEM} lists the detailed quantitative comparison results by using add, concatenate and subtract operator, respectively. 
The results show that the subtract operator, \emph{i.e.}, subtract positional encoding from the query feature, achieves the best performance in both IDF1 and MOTA.
Hence, we use the subtract operator in our final configuration.

\begin{table}[tp]
\begin{center}
\caption{Results for different fusion strategies in decoder.}
\label{table:abs_DEM}
\begin{tabular}{p{2.8cm} c c }
\hline
Fusion Strategy & IDF1 & MOTA \\
\hline
Add &  55.9 & 61.0  \\
Concatenate & 61.6 & 61.2 \\ 
Subtract & 69.2 & 63.2  \\
\hline
\end{tabular}
\end{center}
\end{table}

\begin{table}[tp]
\begin{center}
\caption{Effect of the number of encoder and decoder layers.}
\label{table:abs_layer}
\begin{tabular}{p{2.8cm}<{\centering}  c c }
\hline
Layer Num & IDF1 & MOTA \\
\hline    
1 &  68.6 & 62.9  \\  
2 &  69.2 & 63.2  \\  
3 &  67.8 & 62.7  \\
4 &  67.3 & 62.7  \\
\hline
\end{tabular}
\end{center}
\end{table}

\begin{table}[!tp]
\begin{center}
\caption{Influence of the length of the temporal window $T$.}
\label{table:history-length}
\begin{tabular}{p{2.8cm}<{\centering} c  c }
\hline
Temporal Window & IDF1 & MOTA \\
\hline    
50 &  68.9 & 63.0  \\
100 &  69.6 & 63.1  \\
150 &  69.2 & 63.2  \\
300 & 68.4 & 62.0 \\
\hline
\end{tabular}
\end{center}
\end{table}

\textbf{Number of Encoder and Decoder Layers.} 
We evaluate the tracking performance over various number of encoder and decoder layers. As shown in Table~\ref{table:abs_layer}, both IDF1 and MOTA improve when the number of layers is increased from 1 to 2. However, the performance decreases when the number of layers increases further. One possible reason is that the training data is not enough to train such a large model. We set the number of layers as 2 in the following experiments.

\textbf{Influence of the Length of the Temporal Window.} 
The length of the temporal window $T$, which determines the maximum frame gap that the tracklets and the detections can be associated, is critical for the final tracking performance. 
Intuitively, increasing the length of the temporal window $T$ allows our method to handle longer occlusions. Hence, one would expect higher values to yield better performance.
We test this hypothesis in Table~\ref{table:history-length} by showing the detailed IDF1 and MOTA scores with respect to $T$. 
As expected, enlarging the temporal window from $T=50$ to $T=150$ improves the overall performance. 
However, the performance decreases when the temporal window is unduly large ($T=300$). One possible reason might be that the reliability of the spatial-temporal modeling decreases with the increase of the temporal window, thus degrading the tracking performance. Hence, we use $T=150$ in our final configuration.

\begin{table*}[tp]
\begin{center}
\caption{Performance comparison with start-of-the art on three MOTChallenge benchmarks. The online methods are indicated with ($\mathbb{O}$). The arrows in the table indicate low or high optimal metric values. Note that we used bold for the best number.}
\label{table:mot_eval}
\begin{tabular}{p{3.5cm} c c c c c c c c c }
\hline
Method & MOTA $\uparrow$ & IDF1$\uparrow$ & HOTA$\uparrow$ & MT $\uparrow$ & ML$\downarrow$ & FP $\downarrow$ & FN $\downarrow$ & IDs $\downarrow$ & Hz $\uparrow$ \\ 
\hline
& & & & MOT16\\
\hline
BLSTM$\textunderscore$MTP$\textunderscore$O~\cite{Kim_2021_CVPR} ($\mathbb{O}$) & 48.3 & 53.5 & 39.7 & 17.0 & 38.7 & 9792 & 83707 & 735 & \textbf{21.0} \\
Tracktor++v2~\cite{Bergmann_2019_ICCV} ($\mathbb{O}$) & 56.2 & 54.9 & 44.6 & 20.7 & 35.8 & \textbf{2394} & 76844 & 1068 & 1.6 \\
MPNTrack~\cite{braso2020learning} & 58.6 & 61.7 & 48.9 & 27.3 & 34.0 & 4949 & 70252 & \textbf{354} & 6.5 \\
LPC~\cite{peng2021lpc} & 58.8 & 67.6 & 51.7 & 27.3 & 35.0 & 6167 & 68432 & 435 & 4.3 \\
Aplift~\cite{HorKai2021} & 61.7 & 66.1 & 51.3 & \textbf{34.3} & 31.2 & 9168 & 60180 & 495 & 0.6 \\
TMOH~\cite{Stadler_2021_CVPR} ($\mathbb{O}$) & 63.2 & 63.5 & 50.7 & 27.0 & 31.0 & 3122 & 63376 & 1486 & 0.7 \\
TranSTAM ($\mathbb{O}$) & \textbf{63.8} & \textbf{70.6} & \textbf{54.7} & 30.3 & \textbf{30.6} & 7412 & \textbf{57975} & 629 & 11.2 \\
\hline
& & & & MOT17\\
\hline
BLSTM$\textunderscore$MTP$\textunderscore$O~\cite{Kim_2021_CVPR} ($\mathbb{O}$) & 51.5 & 54.9 & 41.3 & 20.4 & 35.5 & 29616 & 241619 & 2566 & \textbf{20.1} \\
Tracktor++v2~\cite{Bergmann_2019_ICCV} ($\mathbb{O}$) & 56.3 & 55.1 & 44.8 & 21.1 & 35.3 & \textbf{8866} & 235449 & 1987 & 1.5 \\
MPNTrack~\cite{braso2020learning} & 58.8 & 61.7 & 49.0 & 28.8 & 33.5 & 17413 & 213594 & 1185 & 6.5 \\
LPC~\cite{peng2021lpc} & 59.0 & 66.8 & 51.5 & 29.9 & 33.9 & 23102 & 206848 & \textbf{1122} & 4.8 \\
Aplift~\cite{HorKai2021} & 60.5 & 65.6 & 51.1 & \textbf{33.9} & 30.9 & 30609 & 190670 & 1709 & 1.8 \\
CenterTrack~\cite{zhou2020tracking} ($\mathbb{O}$) & 61.5 & 59.6 & 48.2 & 26.4 & 31.9 & 14076 & 200672 & 2583 & 17.0 \\
TMOH~\cite{Stadler_2021_CVPR} ($\mathbb{O}$) & 62.1 & 62.8 & 50.4 & 26.9 & 31.4 & 10951 & 201195 & 1897 & 0.7 \\
TranSTAM ($\mathbb{O}$) & \textbf{63.0} & \textbf{69.9} & \textbf{54.6} & 30.4 & \textbf{30.6} & 23022 & \textbf{183659} & 1842 & 10.8 \\
\hline
& & & & MOT20\\
\hline
Tracktor++v2~\cite{Bergmann_2019_ICCV} ($\mathbb{O}$) & 52.6 & 52.7 & 42.1 & 29.4 & 26.7 & \textbf{6930} & 236680 & 1648 & 1.2 \\
LPC~\cite{peng2021lpc} & 56.3 & 62.5 & 49.0 & 34.1 & 25.2 & 11726 & 213056 & 1562 & 0.7\\
MPNTrack~\cite{braso2020learning} & 57.6 & 59.1 & 46.8 & 38.2 & 22.5 & 16953 & 201384 & \textbf{1210} & \textbf{6.5} \\
Aplift~\cite{HorKai2021} & 58.9 & 56.5 & 46.6 & 41.3 & 21.3 & 17739 & 192736 & 2241 & 0.4 \\
TMOH~\cite{Stadler_2021_CVPR} ($\mathbb{O}$) & \textbf{60.1} & 61.2 & 48.9 & \textbf{46.7} & \textbf{17.8} & 38043 & 165899 & 2342 & 0.6 \\
TranSTAM ($\mathbb{O}$) & \textbf{60.1} & \textbf{66.7} & \textbf{51.7} & 46.5 & \textbf{17.8} & 37657 & \textbf{165866} & 2926 & 4.0 \\
\hline
\end{tabular}
\end{center}
\end{table*}

\subsection{Benchmark Evaluation}
We report the quantitative results obtained by our TranSTAM on the test sets of three MOTChallenge benchmarks MOT16, MOT17 and MOT20 in Table~\ref{table:mot_eval}, and follow the standard evaluation practice to compare it to methods that are officially published and peer reviewed on the three benchmarks. 
It should be noticed that both \textit{online} and \textit{offline} approaches are presented, and the online methods are indicated with ($\mathbb{O}$).
All the results are sorted with ascending MOTA.
As shown in Table~\ref{table:mot_eval}, our approach surpasses the state-of-the-art online methods on all evaluated benchmarks by a large margin, improving especially the IDF1 measure by 7.1, 7.1, and 5.5 percentage points respectively, and the HOTA measure by 4.0, 4.2, and 2.8 percentage points respectively. 
It demonstrates that our method can achieve strong performance in identity preservation.
It is worth noting that our method even achieves better performance than all the offline methods.
Moreover, owing to the concise network and no complex post-processing, our approach is faster than most of the online and offline methods.

\begin{table*}[tp]
\begin{center}
\caption{Performance comparison with TransMOT on MOT17 benchmark. The arrows in the tables indicate low or high optimal metric values. Note that we used bold for the best number.}
\label{table:comparison_with_tra}
\begin{tabular}{p{2.0cm} c c c c c c c c c }
\hline
Method & MOTA $\uparrow$ & IDF1$\uparrow$ & HOTA$\uparrow$ & MT $\uparrow$ & ML$\downarrow$ & FP $\downarrow$ & FN $\downarrow$ & IDs $\downarrow$ & Hz $\uparrow$ \\ 
\hline
& & & & Overall\\
\hline
TransMOT~\cite{chu2021transmot} & \textbf{76.7} & \textbf{75.1} & \textbf{61.7} & 1200 & 387 & 36231 & \textbf{93150} & \textbf{2346} & 1.1 \\
TranSTAM & 76.1 & 74.0 & 60.7 & \textbf{1203} & 387 & \textbf{36213} & 93243 & 5343 & \textbf{10.4} \\
\hline
& & & & MOT17-01\\
\hline
TransMOT~\cite{chu2021transmot} & \textbf{65.9}  & 66.1  & 52.6 & 10 & 3 & 303  & 1876 & \textbf{18} & - \\
TranSTAM & 65.6 & \textbf{77.3} & \textbf{57.1} & 10 & 3 & \textbf{295} & 1876 & 46 & - \\
\hline
& & & & MOT17-03\\
\hline
TransMOT~\cite{chu2021transmot} & \textbf{90.6} & \textbf{87.1} & \textbf{71.2} & \textbf{134} & 3 & \textbf{2820} & 6898 & \textbf{86} & - \\
TranSTAM & 90.5  & 84.2  & 69.8 & 133 & 3 & 2843 & \textbf{6924} & 222 & - \\
\hline
& & & & MOT17-06\\
\hline
TransMOT~\cite{chu2021transmot} & \textbf{58.5} & \textbf{61.4} & \textbf{50.3} & 100 & 46 & 1722 & 3016 & \textbf{155} & - \\
TranSTAM & 57.1 & 50.1 & 42.6 & \textbf{101} & 46 & \textbf{1718} & \textbf{3012} & 327 & - \\
\hline
& & & & MOT17-07\\
\hline
TransMOT~\cite{chu2021transmot} & \textbf{67.8} & 56.2 & 47.6 & 32 & 2 & 1783 & 3539 & \textbf{120} & - \\
TranSTAM & 67.1 & \textbf{67.9} & \textbf{53.0} & 32 & 2 & \textbf{1781} & \textbf{3537} & 232 & - \\
\hline
& & & & MOT17-08\\
\hline
TransMOT~\cite{chu2021transmot} & \textbf{55.7} & 47.9 & \textbf{41.9} & \textbf{28} & 9 & 1847 & \textbf{7297} & \textbf{220} & - \\
TranSTAM & 53.9 & \textbf{50.1} & 41.7 & 27 & 9 & \textbf{1835} & 7310 & 595 & - \\
\hline
& & & & MOT17-12\\
\hline
TransMOT~\cite{chu2021transmot} & \textbf{51.4} & \textbf{66.1} & \textbf{53.1} & 45 & 24 & 1791 & 2383 & \textbf{35} & - \\
TranSTAM & 51.0 & 62.3 & 51.8 & \textbf{46} & 24 & 1791 & 2383 & 70 & - \\
\hline
& & & & MOT17-14\\
\hline
TransMOT~\cite{chu2021transmot} & \textbf{56.7} & \textbf{66.6} & \textbf{49.5} & 51 & 42 & 1811 & 6041 & \textbf{148} & - \\
TranSTAM & 56.0 & 63.6 & 47.7 & \textbf{52} & 42 & \textbf{1808} & \textbf{6039} & 289 & - \\
\hline
\end{tabular}
\end{center}
\end{table*}

\subsection{Additional Comparison with TransMOT }\label{section:additional_experiment}
It should be noticed that there is no peer reviewed Transformer-based MOT methods.
In this subsection, we provide an extended comparison of our method with TransMOT~\cite{chu2021transmot} which is a top-performing Transformer-based MOT method.
We directly use the TransMOT's results that are officially published on the MOTChallenge benchmark.
Note that TransMOT uses private detections, we adopt the same set of detections as TransMOT for a fair comparison. 

The results are summarized in Table~\ref{table:comparison_with_tra}.
Overall, TransMOT outperforms our method by 0.6 percentage points in MOTA measure and 1.1 percentage points in IDF1 measure.
If further refined to each video sequence, our method outperforms TransMOT by a large margin in terms of IDF1 score on MOT17-01, MOT17-07, MOT17-08, while getting worse IDF1 score than TransMOT on MOT17-06, MOT17-12, MOT17-14.
It should be noticed that MOT17-01, MOT17-07 and MOT17-08 are recorded by static cameras, while MOT17-06, MOT17-12 and MOT17-14 are recorded by moving cameras.
It demonstrates that our method may achieve better performance in identity preservation than TransMOT on static video sequences, while getting worse performance on moving video sequences.
One possible reason is that the effectiveness of the relative spatial-temporal positional encoding (RSTPE) decreases in moving video sequences.
In TransMOT, there are a few additional modules needed for data association, such as the Kalman-Filter based motion predictor, the long-term occlusion and duplicated detection handling module.
It will increase the computational cost. 
In contrast, our method does not require any additional modules, hence being up to one order of magnitude faster than TransMOT, i.e., 10.4Hz vs 1.1Hz. 

\section{Conclusion}\label{section:conclusion}
In this paper, we propose a novel Transformer-based method named TransSTAM for online MOT. 
The key innovations of TransSTAM are three simple yet effective positional encoding methods. TransSTAM has two major advantages: (1) Its architecture is compact and is computationally efficient; (2) It can effectively learn spatial-temporal and appearance features jointly within single model, hence achieving better tracking accuracy.
The superiority of the proposed TransSTAM is shown on three popular benchmarks, where we achieve state-of-the-art results.

\clearpage
%
%
\bibliographystyle{splncs04}
\bibliography{egbib}
\end{document}


\pagestyle{headings}
\mainmatter
\def\ECCVSubNumber{251}  

\title{Joint Spatial-Temporal and Appearance Modeling with Transformer for Multiple Object Tracking\\ (Supplementary Material)} 

\titlerunning{ECCV-22 submission ID \ECCVSubNumber} 
\authorrunning{ECCV-22 submission ID \ECCVSubNumber} 
\author{Anonymous ECCV submission}
\institute{Paper ID \ECCVSubNumber}

\maketitle
\thispagestyle{empty}
\section{Detailed Architecture}\label{section:supplementary_material_A}
The detailed architecture of TranSTAM is given in Table~\ref{table:achitecture}.

\begin{table}[htp]
\begin{center}
\caption{Architectures for TranSTAM. $H_i$ and $D_i$ is the number of heads and embedding feature dimension in the $i$th MHSA module. $R_i$ is the feature dimension expansion ratio in the $i$th MLP layer. $L_i$ is the feature dimension in the $i$th Linear layer.}
\label{table:achitecture}
\begin{tabular}{l | c | c| c  }
\hline
& Layer Name & Tracklet Path & Detection Path \\ 
\hline
\multirow{Proj.} & Linear & \multicolumn{2}{c}{$L_1 = 256$} \\
      & MLP & \multicolumn{2}{c}{$L_1=256, N = 3$}\\
\hline
\multirow{Encoder \& Decoder} &  MHSA & \multirow{
$\left[\begin{array}{c}
  H_1 = 8, D_1 = 256  \\
  \\
     R_1 = 4 \\
     (L_1=5) \times 8  \\
\end{array}\right]\times 2$} & \multirow{
$\left[\begin{array}{c}
  H_1 = 8, D_1 = 256  \\
  H_2 = 8, D_2 = 256 \\
  R_1 = 4         \\
  (L_1=5, L_2 = 5) \times 8 \\
\end{array}\right] \times 2 $} \\
& MHSA &  & \\
& MLP  &   & \\
& Linear & & \\
\hline
Head & MLP &  \multicolumn{2}{c}{$R_2 = 4$}\\
\hline
\multicolumn{2}{c}{Params} & \multicolumn{2}{c}{10.07M}\\
\hline
\end{tabular}
\end{center}
\end{table}

\clearpage
%
%
\bibliographystyle{splncs04}
\bibliography{egbib}